\documentclass{article}
\usepackage{ijcai18}

\usepackage{times}
\usepackage{xcolor}
\usepackage{soul}
\usepackage[utf8]{inputenc}
\usepackage[small]{caption}

\usepackage{graphicx}
\graphicspath{ {../../appendix/material/} }

\usepackage{boxedminipage}

\usepackage{amsmath}
\usepackage{subfigure}
\usepackage{amsfonts,amssymb}

\usepackage{nicefrac}       

\usepackage{color}
\usepackage[linesnumbered,ruled,vlined]{algorithm2e}
\newtheorem{theorem}{Theorem}[section]

\newtheorem{proposition}[theorem]{Proposition}

\newtheorem{corollary}[theorem]{Corollary}

\title{Differential Equations for Modeling Asynchronous Algorithms}

\author{
Li He$^{1,2}$\thanks{This work was done when the authors were visiting Microsoft Research Asia.},
Qi Meng$^{3*}$,
Wei Chen$^4$,
Zhi-Ming Ma$^{1,2}$,
Tie-Yan Liu$^4$
\\
$^1$ University of Chinese Academy of Sciences, heli@amss.ac.cn \\
$^2$ Academy of Mathematics and Systems Science, Chinese Academy of Sciences, mazm@amt.ac.cn\\
$^3$ Peking University, qimeng13@pku.edu.cn\\
$^4$ Microsoft Research, \{wche, tie-yan.liu\}@microsoft.com 
}

\begin{document}

\maketitle

\begin{abstract}
 Asynchronous stochastic gradient descent (ASGD) is a popular parallel optimization algorithm in machine learning. Most theoretical analysis on ASGD take a discrete view and prove upper bounds for their convergence rates. However, the discrete view has its intrinsic limitations: there is no characterization of the optimization path and the proof techniques are induction-based and thus usually complicated. Inspired by the recent successful adoptions of stochastic differential equations (SDE) to the theoretical analysis of SGD, in this paper, we study the continuous approximation of ASGD by using stochastic differential delay equations (SDDE). We introduce the approximation method and study the approximation error. Then we conduct theoretical analysis on the convergence rates of ASGD algorithm based on the continuous approximation. There are two methods: \emph{moment estimation} and \emph{energy function minimization} can be used to analyze the convergence rates. Moment estimation depends on the specific form of the loss function, while energy function minimization only leverages the convex property of the loss function, and does not depend on its specific form. In addition to the convergence analysis, the continuous view also helps us derive better convergence rates. All of this clearly shows the advantage of taking the continuous view in gradient descent algorithms.

\end{abstract}

\section{Introduction}

Asynchronous stochastic gradient descent (ASGD) is a popular parallel optimization algorithm in machine learning \cite{langford2009slow,agarwal2011distributed,recht2011hogwild,Lian2015Asynchronous,zhang2015deep}. It has been broadly used in solving deep neural network and received many successes in practice recently, which significantly reduce the communication overhead by avoiding idleness. The main issue with asynchronous algorithms lies on using delayed stochastic gradient information. Suppose $\{(a_1,b_1),\cdots,(a_i,b_i),\cdots,(a_n,b_n)\}$ is the training data set, where the input vector $a_i\in\mathbb{R}^d$ and the output $b_i\in\mathbb{R}$. Supervised machine learning algorithms aim to minimize the empirical risk, i.e.,
{\small\begin{equation}\label{loss function}
	\min_{x}F(x)=\frac{1}{n}\sum_{i=1}^nf_i(x),
	\end{equation}}where $f_i(x)=l(x;a_i,b_i)$ is the loss of model $x$ for the training instance $(a_i,b_i)$. ASGD uses multiple threads or local workers to calculate the gradient of the loss function using a mini-batch of training instances. Then the local worker push the gradient to a master and pull the latest parameter from the master. The master uses the received gradient to update the parameter $x$. Because there is no synchronization between the local workers, the gradient which the master received may be a delayed information for the parameter \cite{agarwal2011distributed,recht2011hogwild,Lian2015Asynchronous}. The update rule for ASGD at iteration $k$ can be described as:
{\small\begin{align}
x_{k+1}=x_{k}-\eta_kg_{M_k}(b,x_{k-l_k}),
\end{align}}where $\eta_k$ is the learning rate, $g_{M_k}(b,x_{k-l_k})$ is the stochastic gradient calculated using the minibatch $M_k$ with minibatch size $b$, $l_k\geq 0$ denotes the stochastic delay. In general, it assumes that $l_k, \forall k$ or $\mathbb{E}l_k$ is upper bounded \cite{agarwal2011distributed,Lian2015Asynchronous}.

The delayed information will influence the convergence rate of the optimization algorithms. Theoretical analyses have been conducted on ASGD for various problem settings, mostly from a discrete view, in which the convergence rate is proved by induction for the sequence of iterative of the optimization algorithm. Then they compare it with the convergence rate of sequential stochastic gradient descent to get the speedup condition.  For example, \cite{recht2011hogwild} shows that for convex problem, if the training data is sparse enough, ASGD can achieve linear speedup. \cite{Lian2015Asynchronous} shows that for nonconvex problems, if the delay is upper bounded by the stochastic sampling variance, ASGD can achieve linear speedup.

However, the discrete view has its intrinsic limitations. (1) It cannot provide an explicit characterization of the optimization path, thus lacks insights about the optimization process. (2) The proofs of optimization algorithms are usually induction-based and somehow unavoidably complicated.

In recent years, researchers study the dynamics of optimization algorithms by taking a continuous view. They derive the corresponding differential equations of the existing discrete-time optimization methods by regarding the optimization rule as the numerical solution of a differential equation. Many works \cite{Raginsky2012Continuous,wibisono2016variational,su2014differential,li2016online,Yang2017The,mandt2016variational,Krichene2015Accelerated} study sequential optimization algorithms using continuous view from various aspects. For example, \cite{Raginsky2012Continuous} used the continuous time analysis to study mirror descent algorithm. In \cite{wibisono2016variational} and \cite{su2014differential}, some accelerated methods were studied by the continuous techniques such as second-order ordinary differential equations. In \cite{li2017dynamics}, the authors developed the method of stochastic modified equations (SME), in which stochastic gradient algorithms are approximated in the weak sense by continuous-time SDE. These works provide a clearer understanding of the dynamic optimization process than the previous works that only take the discrete view, thus have the potential in addressing the aforementioned limitations of the discrete view. However, most of the above works are focusing on sequential algorithms, and using continuous techniques to study asynchronous algorithms has not been studied yet.

Inspired by these works, in this paper, we study the continuous approximation of asynchronous stochastic gradient descent. First, we propose a procedure to approximate asynchronous stochastic gradient based algorithms by using stochastic differential delay equations. Then we analyze the approximation error and show that the approximation error is related to the number of iteration $K$ and mini-batch size $b$.

Second, within this approximation, we study the convergence rates by using continuous techniques and show the following results.
(1) For the linear regression problem which is solved by SGD with constant learning rate, we can attain full information of the optimization path, including the first and second-order moments.
(2) For SDDEs, although it is hard to obtain full information, we can still analyze the optimization path by using \emph{moment estimation} and \emph{energy function minimization}. Moment estimation depends on the specific form of the loss function but has nothing to do with its convexity property, whereas energy function minimization leverages the convexity property but does not depend on its specific form. (3) By using these two techniques, in addition to characterizing the optimization path, we also get the convergence rates of optimization algorithms, with a much simpler proof than that from discrete view. Specifically, we prove a tighter convergence rate for ASGD than the other existing results \cite{zheng2016asynchronous,recht2011hogwild}.

All of these results clearly demonstrate the advantages of taking the continuous view in analyzing ASGD algorithm.

\section{Backgrounds}\label{backg}
In this section, we briefly review the basic settings of asynchronous stochastic gradient descent algorithm, and introduce the stochastic differential delay equations.

\subsection{Basic Settings of ASGD}

\noindent{\bf Asynchronous Stochastic Gradient Descent} is an efficient parallel algorithm. Local workers or threads do not need to wait for others to do synchronization, thus the model is updated faster compared with the synchronous SGD. ASGD updates the parameter using a delayed gradient and has been proved achieving linear or sub-linear speedup under certain conditions such as the training data are sparse or the delay can be upper bounded \cite{agarwal2011distributed,recht2011hogwild,Lian2015Asynchronous}.
In this paper, we consider the update rule of ASGD under the consistent read setting \cite{Lian2015Asynchronous}. Let $M_k$ be the index of a mini-batch with size $b$ and $\eta_k$ be the learning rate for $k$-th iteration. We denote {\small $g_{M_k}(b,x_k)=\frac{1}{b}\sum_{i\in M_k}{\nabla f_i(x_k)}$} as the averaged gradient calculated by a mini-batch. The update rule of ASGD is
{\small
	\begin{equation} x_{k+1}=x_k-\eta_k g_{M_k}(b,x_{k-l_k}),
\end{equation}}where $l_k$ is a random delay satisfying $0\leq l_k\leq l$.\footnote{ASGD under inconsistent read setting will be studied as the future work.} Sequential SGD is a special case of ASGD with $l_k=0, \forall k$.

\subsection{Stochastic Differential Delay Equation}

Stochastic differential delay equation (SDDE) for a $d$-dimensional continuous process $X=(X(t))_{t\ge0}$  is a natural generalization of stochastic differential equation (SDE) by allowing the coefficients depending on values in the past, which has been studied by researchers \cite{mao2007stochastic,bao2016asymptotic}. We consider the following form:
{\small\begin{equation*}
	dX(t)=g_1(t, X_t)dt+g_2(t, X_t)dB(t),\ t\in[t_0, T],
	\end{equation*}}where {\small ${X}_t=\{X(t-\theta(t)): 0\le \theta(t) \le \tau\}$} is the segment (or the functional solution), $\theta(t)$ is a random variable describing the random delay and $\tau$ is its upper bound. The symbol
{\small $B(t)$} denotes $r$-dimensional Brownian motion  \cite{durrett2010probability}. Denote by $C([-\tau,0];\mathbb{R}^d)$ the family of continuous functions $\xi$ from $[-\tau,0]$ to $\mathbb{R}^d$ with the norm {\small$\|\xi\|=sup_{s\in[-\tau,0]}\|\xi(s)\|$}. Then $X_t$ is regarded as a $C([-\tau,0];\mathbb{R}^d)$-valued stochastic process. The vector $g_1 \in \mathbb{R}^d$ and the matrix $g_2 \in \mathbb{R}^{d}\times\mathbb{R}^{r}$ are appropriate functions. For the sake of simplicity, we consider the case of $d=r$ in this paper. SDE can be regarded as a special case of SDDE with $\theta(t)=0, \forall t$. 

\section{Continuous Approximation of Asynchronous Optimization Methods}\label{rep}

In this section, we first propose the continuous approximation of ASGD by using SDDE and describe the approximation procedure in detail. Then we analyze the approximation error for the continuous approximation of ASGD.

First, we introduce some notations. We relate the discrete iterations with continuous stochastic process by introducing the ansatz $x_k\approx X(k\delta)$ where $X(t):\mathbb{R}^+\cup\{0\}\rightarrow \mathbb{R}^d$ is a continuous function of $t$. Given $t\in[0,T]$, we have {\small$T=\delta K$}, where $K$ is the total number of discrete iterations and $\delta$ is the interval to do discretization of $X(t)$ which we call the time step. The discrete index of iterations and continuous index of time have the relation:  {\small $X(t+\delta)\approx x_{\nicefrac{(t+\delta)}{\delta}}:=x_{k+1}$}, {\small $X(t)\approx x_{\nicefrac{t}{\delta}}:=x_k$}. We define the covariance matrix of $\nabla f_i(x_k)$ as
{\small
\begin{equation*}
\Sigma(x_k):=\mathbb{E}(\nabla f_i(x_k)-\mathbb{E}\nabla f_i (x_k))(\nabla f_i(x_k)-\mathbb{E}\nabla f_i(x_k))^T.
\end{equation*}}Since $g_{M_k}(b,x_k)$ is a sum of $b$ i.i.d random vectors $\nabla f_i(x_k)$, then the covariance matrix of $g_{M_k}(b,x_k)$ is $\frac{\Sigma(x_k)}{b}$ \cite{Balles2016Coupling,li2017dynamics}.

\subsection{Asynchronous Gradient Methods}\label{delay}
Most existing approximations are built for sequential algorithms \cite{li2017dynamics,mandt2016variational}, while asynchronous gradient methods are widely used due to its efficiency in utilizing multiple computational nodes \cite{recht2011hogwild,agarwal2011distributed,dean2012large,zhang2015deep,zheng2016asynchronous}.
Therefore, in this section, we describe how to approximate the update rule of ASGD using SDDEs.

Consider the following asynchronous gradient methods
{\small\begin{eqnarray}\label{eq26}
x_{k+1}=x_k-\eta u_k g_{M_k}(b,x_{k-l_k})\label{ASGD},
\end{eqnarray}}where $\eta$ is a deterministic learning rate and $u_k\in[0,1]$ is an adjustment function of $k$. The symbol $l_k$ denotes the random delay and $g_{M_k}(b,x_{k-l_k})$ denotes the delayed information such as delayed gradient.

We first give the following proposition (Theorem 3.9.6, \cite{durrett2010probability}, \cite{Balles2016Coupling}), on which our transformation and approximation error analysis are based.

\begin{proposition}\label{prop}
The $\nabla F(x_{k-l_k})-g_{M_k}(b,x_{k-l_k})$ converges weakly to a multivariate normal distribution with mean $\vec{0}$ and covariance matrix $\nicefrac{\Sigma(x_{k-l_k})}{b}$.
\end{proposition}
Moreover, we assume positive definiteness and a square root decomposition of the covariance matrix of $\nicefrac{\Sigma(x_{k-l_k})}{b}$, i.e., {\small$\frac{\Sigma(x_{k-l_k})}{b}=\sigma(x_{k-l_k})\sigma(x_{k-l_k})^T$}.
Reformulating Eq.(\ref{ASGD}) as:
{\small\begin{align*}
x_{k+1}= &x_k-\eta u_k \nabla F(x_{k-l_k})\\
&+\eta u_k (\nabla F(x_{k-l_k})-g_{M_k}(b,x_{k-l_k})).
\end{align*}}Let $z_k$s be i.i.d random vectors which follow standard normal distribution $N(\vec{0},I_{d\times d})$. Thus, we can use $\sigma(x_{k-l_k}) z_k$ to approximate the noise term (by Central Limit Theorem). Therefore, the above equation can be approximated by
{\small
\begin{equation}\label{approx}
x_{k+1}=x_k-\eta u_k\nabla F(x_{k-l_k})+\eta u_k\sigma(x_{k-l_k}) z_k.
\end{equation}
}Choosing a precision $\delta$, we can view the Eq.(\ref{approx}) as an Euler-Maruyama approximation of the following SDDE \cite{mao2007stochastic}:
{\small\begin{align}\label{sdde}		
\mathrm{d}X(t)=
&-\frac{\eta}{\delta}U(t)\nabla F(X(t-\theta(t)))\mathrm{d}t\\\nonumber
&+\frac{\eta}{\delta}\sqrt{\delta}U(t)\sigma(X(t-\theta(t)))\mathrm{d}B(t),
\end{align}}where $\theta(t)=l_k\delta$. We assume that $0\leq l_k\leq l$ and denote $\tau:=\l\delta$. In the following, we give several possible choices for the time step under different cases. (a) For ASGD with constant learning rate $\eta=\eta_0$ and $u_k=1$, we choose the $\delta=\eta_0$. (b) For decreasing learning rate $\eta=\eta_0$ and $u_k=\frac{1}{k}$, we have $U(t)=\frac{\delta}{t}$. We can choose any precision. (c) For $\eta=\eta_0$ and $u_k=\frac{1}{\sqrt{k}}$, since $U(t)=\frac{\sqrt{\delta}}{\sqrt{t}}$, we choose $\delta=\eta_0$.

In summary, we used SDDE to approximate ASGD through Eq.(\ref{eq26}), Eq.(\ref{approx}) and Eq.(\ref{sdde}). Many asynchronous gradient methods can be represented by Eq.(\ref{eq26}), such as ASGD, asynchronous SVRG, etc.
	
Using above similar technique, Nesterov's accelerated and momentum stochastic gradient methods can be transformed into two stage SDEs. Besides, the continuous approximation can be described by a second-order SDE. We omit the explicit transform because it is not the main case which will be studied in this paper. For more details, it can be referred in \cite{su2014differential,li2017dynamics}.

\subsection{Approximation Error}
In this subsection, we study the approximation error between $X(t_k)$,
produced by the continuous process and $x_k$ which is produced by the discrete process. We applied Central Limit Theorem and Euler-Maruyama approximation of stochastic delay differential equation in analyzing approximation error. Now we give the following Theorem \ref{thm2}-\ref{thm3.3} and proof sketches, which show the approximation error between the update produced by ASGD (Eq.(\ref{eq26})) and its corresponding SDDE (Eq.(\ref{sdde})) in different cases.

\begin{theorem}\label{thm2}
	Assume that $\nabla F(x)$ is $L$-Lipschitz continuous, and $b$ is the mini-batch size. Consider delay $l_k$ with upper bound $l$ and $u_k=1$ for $k=1,\dots,K$. If we consider ASGD with $\delta=\eta$, $\eta=\frac{1}{K}$ and $\eta L(l+1)\leq 1$, we have
{\small\begin{equation*}
\mathbb{E}\|X(t_K)-x_K\| \leq C_1(\frac{1}{\sqrt{K}}+\frac{l}{K^{3/2}})+C_2\frac{e^{L(l+1)}}{\sqrt{b}(l+1)},
\end{equation*}
}where $C_1, C_2$ are constants.	
\end{theorem}

\noindent{\textit {Proof sketch:}}
(1) For SDDE with $U(t)=1$, the Euler-Maruyama scheme with $\delta=\eta$ is
{\small
\begin{align*}
\tilde{x}_{k+1}=\tilde{x}_k-\eta\nabla F(\tilde{x}_{k-l_k})+\sqrt{\eta}(B(t_{k+1})-B(t_k))\sigma(\tilde{x}_{k-l_k}),
\end{align*}}where $B(t_{k+1})-B(t_k)$ is usually modeled by $\sqrt{\eta}z_k$. It is well-known that the Euler-Maruyama scheme is strongly convergent with order $\frac{1}{2}$. According to (Theorem 5.5.5 \cite{mao2007stochastic}), we have 
{\small	$\mathbb{E}(\|X(t_k)-\tilde{x}_k\|)\leq\tilde{C}_1\eta^{3/2}(k+l)e^{2(\eta k)^2}.$} Let $\eta=\frac{1}{K}$, we can obtain {\small	$$\mathbb{E}(\|X(t_K)-\tilde{x}_K\|)\leq C_1(\frac{1}{\sqrt{K}}+\frac{l}{K^{3/2}}).$$}(2) Next, let us consider the relationship between $x_{k+1}$ and $\tilde{x}_{k+1}$. 
By the central limit theorem (CLT), we can get that $\nabla F(x_{k-l_k})-g_{M_k}(b,x_{k-l_k})$ converges to $\sigma(x_{k-l_k}) z_k$ in distribution with the increasing minibatch size $b$. Assume that
$\nabla F(x_k)$ is $L$-Lipschitz continuous and using Theorem 3.4.9 in \cite{durrett2010probability}, the gap between their distribution functions goes to zero at rate $b^{-\frac{1}{2}}$. Since ASGD uses a delayed gradient, it will cause the mismatch between $x_k$ and $g_{M_k}(b,x_{k-l_k})$ when we expand the series. We have
{\small
\begin{align*}
&\mathbb{E}\|x_{k+1}-\tilde{x}_{k+1}\|	\\
=&\mathbb{E}\|x_{k}-\tilde{x}_{k}-\eta(\nabla F(x_{k-l_k})- \nabla F(\tilde{x}_{k-l_k}))\\	
&-\eta[\sigma(\tilde{x}_{k-l_k})z_k-\nabla F(x_{k-l_k})+g_{M_k}(b,x_{k-l_k})]\|\\
\leq & \mathbb{E}\|x_{k}-\tilde{x}_{k}-\eta(\nabla F(x_{k-l_k})- \nabla F(\tilde{x}_{k-l_k}))\|\\
&+ \mathbb{E}\|\eta[\sigma(\tilde{x}_{k-l_k}) z_k-\nabla F(x_{k-l_k})+g_{M_k}(b,x_{k-l_k})]\|.
\end{align*}}Denote
{\small $\mathbb{E}\|\eta[\sigma(\tilde{x}_{k-l_k}) z_k-\nabla F(x_{k-l_k})+g_{M_k}(b,x_{k-l_k})]\|$} as $\Phi(k)$, then we have
{\small
\begin{align*}
\Phi(k)\leq &  \eta (\mathbb{E}\|\sigma(\tilde{x}_{k-l_k})) z_k\|^2)^{\frac{1}{2}}\\
&+ \eta(\mathbb{E}\|g_{M_k}(b,x_{k-l_k})-\nabla F(x_{k-l_k})\|^2)^{\frac{1}{2}}\\
=& \eta \sqrt{\frac{tr(\Sigma(\tilde{x}_{k-l_k}))}{b}}+ \eta \sqrt{\frac{tr(\Sigma(x_{k-l_k}))}{b}},
\end{align*}
}where $tr(\Sigma)$ denotes the trace of the matrix and we derived the last step by simple matrix computations.

\noindent Denote $\frac{\eta \Phi}{\sqrt{b}}=sup_{s=0}^{k}\Phi(s)$ as an upper bound. Let  $A_{k}=\mathbb{E}\|x_{k}-\tilde{x}_{k}\|$ and $A_0=0$. We consider a probability $q_j$ is assumed for each $l_k=j,\  j=0,1, \cdots, l$. \footnote{It should be noted that for $k\leq l$, the random delay $l_k$ only take values from the set $\{0,1,\cdots,k\}$.}
Taking expectation and using upper bound, it becomes
{\small
\begin{align*}
A_{k}&\leq A_{k-1}+\eta L\sum_{j=1}^{l+1} q_j A_{k-j}+\frac{\eta \Phi}{\sqrt{b}}\\
&\leq A_{k-1}+\eta L\sum_{j=1}^{l+1}  A_{k-j}+\frac{\eta \Phi}{\sqrt{b}}.
\end{align*}
}Thus we have
{\small$
A_{k}\leq \eta L \sum_{j=1}^{l+1} (A_{1-j} +\dots+A_{k-j})+k\frac{\eta \Phi}{\sqrt{b}}
\leq \eta L (l+1)(A_0+A_1+\dots+A_{k-1})+k\frac{\eta \Phi}{\sqrt{b}},
$
}where {\small $A_0=0, A_1=\eta L (l+1)A_0+\frac{\eta \Phi}{\sqrt{b}}$}. Using this recursion, we obtain
{\small
\begin{align*}
& A_{k}
\leq k \frac{\eta \Phi}{\sqrt{b}}+\eta L (l+1) \frac{\eta \Phi}{\sqrt{b}}\sum_{j=0}^{k-2} (1+\eta L (l+1))^j(k-j-1)\\
&\leq k \frac{\eta \Phi}{\sqrt{b}}+\eta L (l+1) \frac{\eta \Phi}{\sqrt{b}}\frac{(1+\eta L (l+1))^{k}}{(\ln (1+\eta L (l+1)))^2}\\
&\leq k \frac{\eta \Phi}{\sqrt{b}}+\eta L (l+1) \frac{\eta \Phi}{\sqrt{b}}\frac{e^{\eta L k(l+1)}}{(\ln (1+\eta L (l+1)))^2}\\
&\leq  k \frac{\eta \Phi}{\sqrt{b}}+\frac{4\frac{\eta \Phi}{\sqrt{b}} e^{\eta L k(l+1)}}{\eta L (l+1)},
\end{align*}
}where we used integration approximation. In the last step, we assumed that $\eta L(l+1)\leq 1$ and used $\frac{\eta L(l+1)}{2}\leq \eta L(l+1)-\frac{(\eta L(l+1))^2}{2} \leq \ln(1+\eta L(l+1)) $. Let $\eta=\frac{1}{K}$, we have
{\small
\begin{equation*}
\mathbb{E}\|X(t_K)-x_K\|
\leq C_1(\frac{1}{\sqrt{K}}+\frac{l}{K^{3/2}})+C_2\frac{e^{L(l+1)}}{\sqrt{b}(l+1)},
\end{equation*}
}where $C_1, C_2$ are constants.     \                     $\Box$

If we consider the upper bound $l=0$, ASGD reduces to SGD. We give the following corollary.

\begin{corollary}\label{corollary1}
	Assume that $\nabla F(x)$ is $L$-Lipschitz continuous, and $b$ is the mini-batch size.
	Consider delay $l_k$ with upper bound $l=0$ and $u_k=1$ for $k=1,\dots,K$. The SGD with $\delta=\eta$, $\eta=\frac{1}{K}$ and $\eta L\leq 1$, we have
	{\small\begin{equation*}
		\mathbb{E}\|X(t_K)-x_K\| \leq C_1\frac{1}{\sqrt{K}}+C_2\frac{e^{L}}{\sqrt{b}},
		\end{equation*}
	}where $C_1, C_2$ are constants.		
\end{corollary}

The above theorem just assumes that the delay can be upper bounded. If we further assume that the randomness of the delay can be neglected, for example, a cyclic delayed update architecture \cite{agarwal2011distributed}, the results can be improved. In this case, we give the following theorem.

\begin{theorem}\label{thm3.3}
	Assume that $\nabla F(x)$ is $L$-Lipschitz continuous, and $b$ is the mini-batch size.
Consider delay $l_k=l$ and $u_k=1$ for $k=1,\dots,K$. If we consider ASGD with $\delta=\eta$, $\eta=\frac{1}{K}$ and $\eta Ll\leq 1$, we have
{\small\begin{equation*}
\mathbb{E}\|X(t_K)-x_K\| \leq C_3(\frac{1}{\sqrt{K}}+\frac{l}{K^{3/2}})+C_4\frac{e^L}{\sqrt{b}}.
\end{equation*}
}where $C_3, C_4$ are constants.	
\end{theorem}

\noindent{\textit {Proof sketch:}}
The analysis between $X(t_k)$ and $\tilde{x}_k$ follows from the proof of Theorem \ref{thm2}. We analyze the relationship between $x_{k}$ and $\tilde{x}_{k}$. Using the above inequality and $A_0=0$, we can get
{\small$$
A_{kl}\leq \eta L\sum_{i=1}^{(k-1)l}A_i+\frac{\eta \Phi}{\sqrt{b}}\cdot kl.
$$}
Furthermore, we have the following recursion relation:
{\small
\begin{equation*}
(A_{(k-1)l}+\cdots+A_{(k-2)l+1})
\leq \eta L l\sum_{i=1}^{(k-2)l}A_i+\frac{\eta \Phi}{\sqrt{b}}\cdot (k-1)l^2.
\end{equation*}}Using that $A_l+\cdots+A_1\leq \frac{2(1+\eta L)^l}{\ln{(1+\eta L)}}$, we conclude that
{\small\begin{align*}	
A_{kl}\leq& \eta L(1+\eta Ll)^{k-1}\cdot\frac{2(1+\eta L)^l\cdot \frac{\eta \Phi}{\sqrt{b}}}{\ln{(1+\eta L)}}\\
&+\frac{\eta^2l^2L\Phi}{\sqrt{b}}\sum_{i=0}^{k-1}(1+\eta Ll)^i(k-1-i)\\
\leq &\eta L(1+\eta Ll)^{k-1}\cdot\frac{2(1+\eta L)^l\cdot \frac{\eta \Phi}{\sqrt{b}}}{\ln{(1+\eta L)}}+\frac{\eta^2l^2L\Phi(1+\eta Ll)^k}{(\ln{(1+\eta Ll)})^2\sqrt{b}}\\
\leq &\frac{2\eta Le^{\eta Llk}\cdot \frac{\eta \Phi}{\sqrt{b}}}{\ln{(1+\eta L)}}+\frac{\eta^2l^2L\Phi e^{\eta Llk}}{(\ln{(1+\eta Ll)})^2\sqrt{b}}.
\end{align*}
}Assume that $\eta L l\leq 1$ and using $\frac{\eta Ll}{2}\leq \ln(1+\eta Ll)$, we obtain
{\small
\begin{align*}
A_{kl}\leq {4e^{\eta Llk}\cdot \frac{\eta \Phi}{\sqrt{b}}}+\frac{4\Phi e^{\eta Llk}}{L\sqrt{b}}\leq C_4\frac{e^{\eta Llk}}{\sqrt{b}}.
\end{align*}} Let $\eta=\frac{1}{K}$, now we have the following bound
{\small\begin{equation*}
\mathbb{E}\|X(t_K)-x_K\| \leq C_3(\frac{1}{\sqrt{K}}+\frac{l}{K^{3/2}})+C_4\frac{e^L}{\sqrt{b}},
\end{equation*}
}where $C_3, C_4$ are constants.                  $\Box$

The Lipschitzs coefficient is relatively a small constant. For example, we have $L\leq 1$ and $L\leq 0.25$ for linear regression and logistic regression, respectively. 
Theorem \ref{thm2} shows that the approximation error of ASGD will be small if the number of iterations $K$ and the minibatch size $b$ are large. The delay $l$ has influence on the two parts. This is consistent with the intuition that delay will cause mismatch of the updates. 
From Theorem \ref{thm3.3}, when the delay equals to a constant, its approximation error is similar to that of SGD since $l$ does not have influence on the part of  $\frac{e^{L}}{\sqrt{b}}$.

\section{Convergence Analysis: Techniques and Examples}\label{PA}
In this section, we show some techniques using SDDE for convergence analysis. Firstly, we introduce the measure for convergence analysis. Taking SDE as a simple example, for fixed $t$, the SDE becomes an ODE after taking expectation of $X(t)$. We can get the optimum $x^*$ by taking limits {\small$\lim_{t\rightarrow \infty}\mathbb{E}X(t)$}. Secondly, we can calculate {\small$\mathbb{E}(X(t)-x^*)$} and {\small$\mathbb{E}||X(t)-\mathbb{E}X(t)||^2$} for each fixed $t$. Combining them we can get the convergence of {\small$\mathbb{E}||X(t)-x^*||^2$}.

Usually, SDEs can be classified into two cases: analytic solution case and non-analytic solution case. An example of SDE with analytic solution is linear regression solved by SGD with constant learning rate. The objective function is
{\small$
F(x)=\frac{1}{2n}\|Ax-B\|^2,$}
with $A=(a_1^T,\dots,a_n^T)^T$ and $B=(b_1,\dots,b_n)^T$, $x\in\mathbb{R}^d, a_i\in\mathbb{R}^d$ and $b_i\in\mathbb{R}$. Denote $\tilde A =\frac{1}{n}A^T A$. The solution of its SDE is an \textit{Ornstein-Uhlenbeck} process \cite{uhlenbeck1930theory}:
{\small
\begin{equation*}
X(t)=x^*+e^{-\tilde{A}t}(X(0)-x^*)+\int_{0}^te^{\tilde{A}(s-t)}\sqrt{\eta}G\mathrm{d}B(s).
\end{equation*}}We can calculate the moments directly, i.e.,
{\small
\begin{align*}		
&\mathbb{E}[X(t)-x^*|X(0)=x_0]=e^{-\tilde{A}t}(x_0-x^*), \\
&\mathbb{E}[\|X(t)-\mathbb{E}X(t)\|^2|X(0)=x_0]\le
\int_{0}^t\|e^{\tilde{A}(s-t)}\sqrt{\eta}G\|^2\mathrm{d}s.
\end{align*}}From the above results, we can get many insights about the dynamics of SGD, such as the oscillation of the sample path which has been discussed in \cite{li2017dynamics}. 

However, SDDEs are more difficult to obtain analytic solution. For non-analytic solution case, it is hard to calculate the exact moments of $X(t)-x^*$, we can use two methods to estimate them to achieve the convergence for ASGD: \emph{moment estimation} and \emph{energy function minimization}.

\subsection{Moment Estimation}
We show the moment estimation technique by taking linear regression as an example. It can also be used for nonconvex loss functions. Before showing the details, we need to guarantee the existence and uniqueness of the solutions for SDDEs. Uniform Lipschitz condition and linear growth condition are two conditions to guarantee the existence and uniqueness of the solutions for SDDE \cite{mao2007stochastic}. We should see that these two conditions have no direct relation with convexity. For our continuous approximation, we do not care the explicit form of {\small$\sigma(X(t-\theta(t)))$} (except for variance reduced techniques are involved such as SVRG \cite{johnson2013accelerating}) and just assume that it has an upper bound thus it is a constant about {\small$X(t-\theta(t))$}. In this case, the diffusion term satisfies linear growth and Lipschitz conditions. We just need to check whether {\small$-\nabla F(X(t-\theta(t)))$} satisfies the two conditions.

We use Theorem 5.2.2 in \cite{mao2007stochastic} to guarantee the existence and uniqueness of the solution to SDDE. 
We consider the ASGD with constant learning rate $\eta$ for the linear regression. The corresponding SDDE is
{\small
\begin{align}\label{linearsdde}
&\mathrm{d}X(t)=-(\tilde A X(t-\theta(t))-\tilde B)\mathrm{d}t+\sqrt{\eta}\sigma(X(t-\theta(t)))\mathrm{d}B(t),\nonumber\\ &X(s)=\xi(s),s\in[-\tau,0],
\end{align}}where $\theta(t)\in[0,\tau]$. Here we consider $\theta(t)=\tau$ and $\tau$ is its upper bound. Following the techniques in \cite{bao2016asymptotic,bao2014ergodicity}, we can bound the first and second moments and give the following theorem.

\begin{theorem}\label{thm4.2}
Define {\small $V=sup\{Re(\beta):\beta\in\mathbb{C},det(\beta I_{d\times d}+\tilde A e^{-\beta \tau})=0\}.$} For any given training error $\epsilon$, let $\eta=\frac{\epsilon}{\tau^2}$. If $V<0$, then for $\lambda\in(0,-V)$, the convergence rate for ASGD for linear regression Eq.(\ref{linearsdde}) by using moment estimation is
{\small
\begin{eqnarray*}
\mathbb{E}\|X(t)-x^*\|^2 \le C_5e^{-2\lambda (t-\tau)}+C_6\frac{\epsilon}{\lambda\tau^2},
\end{eqnarray*}}where $C_5$ and $C_6$ are constants. \footnote{The coefficients are related to the root of characteristic function of the SDDE.}
\end{theorem}

\noindent\textit{Proof sketch:}
Through some lengthy derivation \cite{bao2016asymptotic}, we have {\small
$\|\mathbb{E}(X(t))-x^*\|\le a_{1}e^{-\lambda t}$} and {\small $\mathbb{E}\|X(t)-\mathbb{E}X(t)\|^2\le a_2(1-e^{-2\lambda t})$} where the coefficient $a_1$ is {\small$c_{\lambda}\|\tilde\xi\|+c_{\lambda}\|\tilde\xi\|\|\tilde A\|(e^{\lambda\tau}-1)\frac{1}{\lambda}$}, $a_2$ is $\frac{c_{\lambda}^2\mathbb{E}\|\sqrt{\eta}\sigma\|^2}{2\lambda}$ and $x^*=\tilde A^{-1}\tilde B$. Moreover, we can obtain
{\small
\begin{equation*} \mathbb{E}\|X(t)-x^*\|^2=\mathbb{E}\|X(t)-\mathbb{E}X(t)\|^2+\|\mathbb{E}X(t)-x^*\|^2.
\end{equation*}}Thus by putting $a_1$ and $a_2$ in it , we have
{\small
\begin{align*}	
&\quad\mathbb{E}\|X(t)-x^*\|^2\leq\\
&\frac{c_{\lambda}^2}{\lambda^2}\left[(\|\tilde\xi\|^2(\|\tilde A\|+\lambda)^2+\lambda\eta\mathbb{E}||\sigma||^2)e^{2\lambda\tau}\right]e^{-2\lambda t}+\frac{c_{\lambda}^2 \mathbb{E}\|\sqrt{\eta}\sigma\|^2}{2\lambda}.
\end{align*}}Putting $\eta=\frac{\epsilon}{\tau^2}$ in the above equation, we can get
{\small
\begin{align*}
&\mathbb{E}\|X(t)-x^*\|^2\\
 \leq&\frac{c_{\lambda}^2\cdot e^{-2\lambda t}}{\lambda^2}\left[(\|\tilde\xi\|^2(\|\tilde A\|+\lambda)^2+\lambda\epsilon\mathbb{E}||\sigma||^2/\tau^2)e^{2\lambda\tau}\right]\\
&+\epsilon c_{\lambda}^2\mathbb{E}\|\sigma\|^2/2\tau^2\lambda \leq C_5e^{-2\lambda (t-\tau)}+C_6\frac{\epsilon}{\lambda\tau^2}. \quad\quad \Box
\end{align*}}{\bf Discussion:} We compare the results with the existing convergence rates of ASGD under consistent read setting. If we don't assume the data are sparse, the number of iterations is no less than $\mathcal{O}(\frac{\tau}{\epsilon})$ to achieve {\small$\mathbb{E}\|x_k-x^*\|^2\leq\epsilon$} \cite{zheng2016asynchronous}. Another well-known theoretical result is $\mathcal{O}(\frac{\log(1/\epsilon)}{\epsilon}\cdot\tau^2)$ if we let the sparse coefficients be $1$ and set $\eta=\frac{\epsilon}{\tau^2}$ \cite{recht2011hogwild} \footnote{Please notice that the theoretical analysis in \cite{recht2011hogwild} is proved under consistent read setting.}. Theorem \ref{thm4.2} shows that we need {\small$ e^{-2\lambda(t-\tau)}\leq\epsilon$} in order to make {\small$\mathbb{E}\|X(t)-x^*\|^2\leq\mathcal{O}(\epsilon)$}.  Thus we need {\small$t\geq \mathcal{O}(\tau+\frac{\log{(1/\epsilon)}}{2\lambda})$}, which is in common faster than the previous two results $\mathcal{O}(\frac{\tau}{\epsilon})$ and $\mathcal{O}(\frac{\log{(1/\epsilon)}}{\epsilon}\cdot\tau^2)$.

\subsection{Energy Function Minimization}
In this section, we show the \emph{energy function minimization}. This technique has been explored in many works \cite{su2014differential,Krichene2015Accelerated,wibisono2016variational}. Firstly, we need to define proper \emph{energy function} for corresponding differential equation. Energy function is related to the measure which is used for the convergence rate of optimization, such as $\|X(t)-x^*\|^2$, $F(X(t))-F(x^*)$ and the expected convergence rate for the optimization algorithms. Suppose that we use ASGD to minimize a strongly convex loss function. We can define the energy function for SDDE as {\small $\mathcal{E}(t)=\frac{t-1}{2}\|X(t)-x^*\|^2-\frac{ (\delta+1)(H+2L^2D^2\tau)\ln{t}}{2\mu^2}$} and calculate $\rm{d}\mathbb{E}\mathcal{E}(t)$ by using It\^o formula. If $\rm{d}\mathbb{E}\mathcal{E}(t)\leq 0$, we can get $\mathbb{E}\mathcal{E}(t)$ is a decreasing function about $t$. Then the convergence rate for SDDE is obtained by using $\mathbb{E}\mathcal{E}(t)\leq \mathcal{E}(t_0)$. Thus, the design for energy function aims to make $\rm{d}\mathbb{E}\mathcal{E}(t)\leq 0$. Theorem \ref{thm4.4} shows the convergence rate for SDDE.

\begin{theorem}\label{thm4.4}
For any {\small$F\in\mathcal{F}_{\infty}$} with smooth coefficient $L$, let $X(t)$ be the unique global solution to (\ref{sdde}) with initial conditions {\small$X(s)=\xi(s),s\in[-\tau,0]$} and {\small$X(t)\in d(x_0, D)$, $\forall t$}. Assume that $F(x)$ is strongly convex about $x$ and {\small$\eta_k=\frac{1}{\mu k}$}. Let $H$ be a constant which satisfies {\small$H\ge\mathbb{E}(Tr(\sigma\sigma^T))+2\mu LD^2$}. For any $t>1$,
{\small
	\begin{equation*}\label{asgd}
	\mathbb{E}\|X(t)-x^*\|^2\leq\frac{(\delta+1)(H+2L^2D^2\tau)\ln{t}}{(t-1)\mu^2}.
	\end{equation*}}
\end{theorem}

\noindent{\textit {Proof sketch:}}
For strongly convex case, $\eta_k$ is set as $\eta_k=\frac{1}{\mu k}$.
Then we have $\eta=1$ and $U(t)=\frac{\delta}{\mu t}$. The SDDE is
{\small
\begin{equation*}
\mathrm{d}X(t)= -\frac{1}{\mu t}\nabla F(X(t-\theta(t)))\mathrm{d}t+\frac{\sqrt{\delta}}{\mu t}\sigma(X(t-\theta(t)))\mathrm{d}B(t).
\end{equation*}} We define the energy function
{\small\begin{equation*}	
\mathcal{E}(t)=\frac{t-1}{2}\|X(t)-x^*\|^2-\frac{(\delta+1)(H+2L^2D^2\tau)\ln{t}}{2\mu^2},
\end{equation*}}where {\small$H\ge\mathbb{E}(Tr(\sigma\sigma^T))+2\mu LD^2$}. 
Using It\^o formula and taking expectation, we can get
{\small
\begin{align*}
&\mathrm{d}\mathbb{E}\mathcal{E}\\	=&\frac{1}{2}\mathbb{E}\|X(t)-x^*\|^2\mathrm{d}t-\frac{1}{\mu}\mathbb{E}\langle\nabla F(X(t-\theta(t))),X(t)-x^*\rangle\mathrm{d}t\\
&+\frac{1}{t\mu}\mathbb{E}\langle\nabla F(X(t-\theta(t))),X(t)-x^*\rangle\mathrm{d}t\\
&+\frac{(t-1)\delta}{2\mu^2 t^2}\mathbb{E}Tr(\sigma\sigma^T)\mathrm{d}t-\frac{ (\delta+1)(H+2L^2D^2\tau)}{2t\mu^2}\mathrm{d}t\\
\leq&\left(\frac{L^2D^2\theta(t)}{\mu^2t}+\frac{LD^2}{t\mu}\right)\mathrm{d}t+\frac{\delta}{2\mu^2 t}\mathbb{E}Tr(\sigma\sigma^T)\mathrm{d}t\\
&-\frac{(\delta+1) (H+2L^2D^2\tau)}{2t\mu^2}\mathrm{d}t \leq 0.
\end{align*}}The first equality comes from It\^o formula. The first inequality applied the positiveness of $\frac{1}{2\mu^2t^2}\mathbb{E}Tr(\sigma\sigma^T)$ and strongly convex assumption. 
Therefore, we have
{\small
\begin{equation*}
\mathbb{E}\|X(t)-x^*\|^2\leq\frac{(\delta+1) (H+2L^2D^2\tau)\ln{t}}{(t-1)\mu^2},
\end{equation*}}where $\tau$ is the upper bound of $\theta(t)$.     \       $\Box$

When we consider the special case of $\tau=0$, ASGD algorithm reduces to SGD algorithm. We can similarly define an energy function {\small$
\mathcal{E}(t)=\frac{t-1}{2}\|X(t)-x^*\|^2-\frac{(\delta+1)H\ln{t}}{2\mu^2}$} and we give the following corollary.

\begin{corollary}\label{cor4.3}
For any {\small$F\in\mathcal{F}_{\infty}$} with smooth coefficient $L$, let $X(t)$ be the unique global solution to {\small\begin{equation}\label{SDE}
\mathrm{d}X(t)=-\frac{\eta}{\delta}U(t)\nabla F(X(t))\mathrm{d}t+\frac{\eta}{\delta}\sqrt{\delta}U(t)\sigma(X(t))\mathrm{d}B(t).
\end{equation}}with initial conditions $X(1)=x_0$ and $X(t)\in d(x_0, D)$, $\forall t$. Assume that $F(x)$ is strongly convex about $x$ and {\small$\eta_k=\frac{1}{\mu k}$}. Let {\small$H\ge\mathbb{E}(Tr(\sigma\sigma^T))+2\mu LD^2$}. For any $t>1$,
{\small
\begin{equation*}\label{converg1}	
\mathbb{E}\|X(t)-x^*\|^2\leq\frac{(\delta+1)H\ln{t}}{(t-1)\mu^2}.
\end{equation*}}
\end{corollary}

\noindent{\bf Discussions:} Since the approximation error is relatively small, the convergence rate of SDDE can be approximated as the result of ASGD. (1) We prove a tighter convergence rate for ASGD than the other existing results.  The result in Theorem \ref{thm4.4} is no slower than the convergence rate {\small$\mathcal{O}(\tau/\epsilon)$} \cite{zheng2016asynchronous}. If {\small$2L^2D^2\tau\leq H$}, the negative influence of $\tau$ can be neglected, which means that it is comparable with the serial SGD and it can achieve linear speedup. Compared with the results in \cite{recht2011hogwild}, the speedup condition does not rely on the sparsity condition. If the stochastic variance {\small$\mathbb{E}(Tr(\sigma\sigma^T))$} is large, the condition {\small$2L^2D^2\tau\leq H$} is easier to be satisfied, which is consistent to the results for non-convex case studied in \cite{Lian2015Asynchronous}.  

(2) Corollary \ref{cor4.3} shows that {\small$\mathbb{E}\|x_k-x^*\|^2\leq\frac{(\delta+1)H\ln(\delta k)}{(\delta k-1)\mu^2}$} since $t=\delta k$. It is comparable with the existing convergence rate of {\small$\frac{4H}{k\mu^2}$} \cite{rakhlin2011making} under the same setting.

\section{Experiments}\label{exp}
In this section, we take {\small $F(x)=\frac{1}{2}(x+1)^2+\frac{1}{2}(x-1)^2$} as a simple example. We compared the discrete optimization algorithms with the Euler-Maruyama schemes of stochastic differential equations. 

\begin{figure}[!h]
	\centering
	\subfigure[]{
		\label{fig1c}
		\includegraphics[width=1.60in]{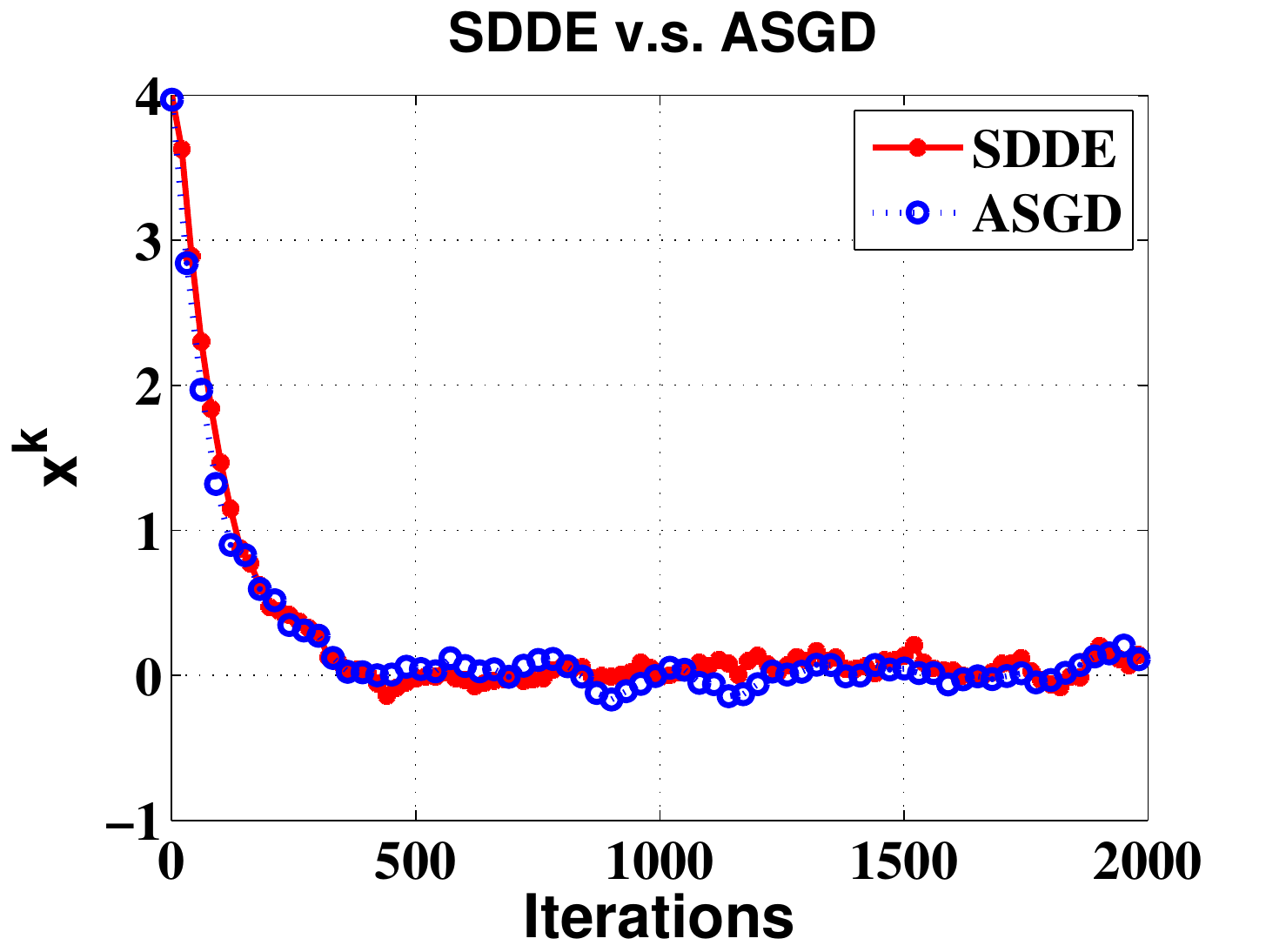}}
    \subfigure[]{
		\label{fig1a}
		\includegraphics[width=1.60in]{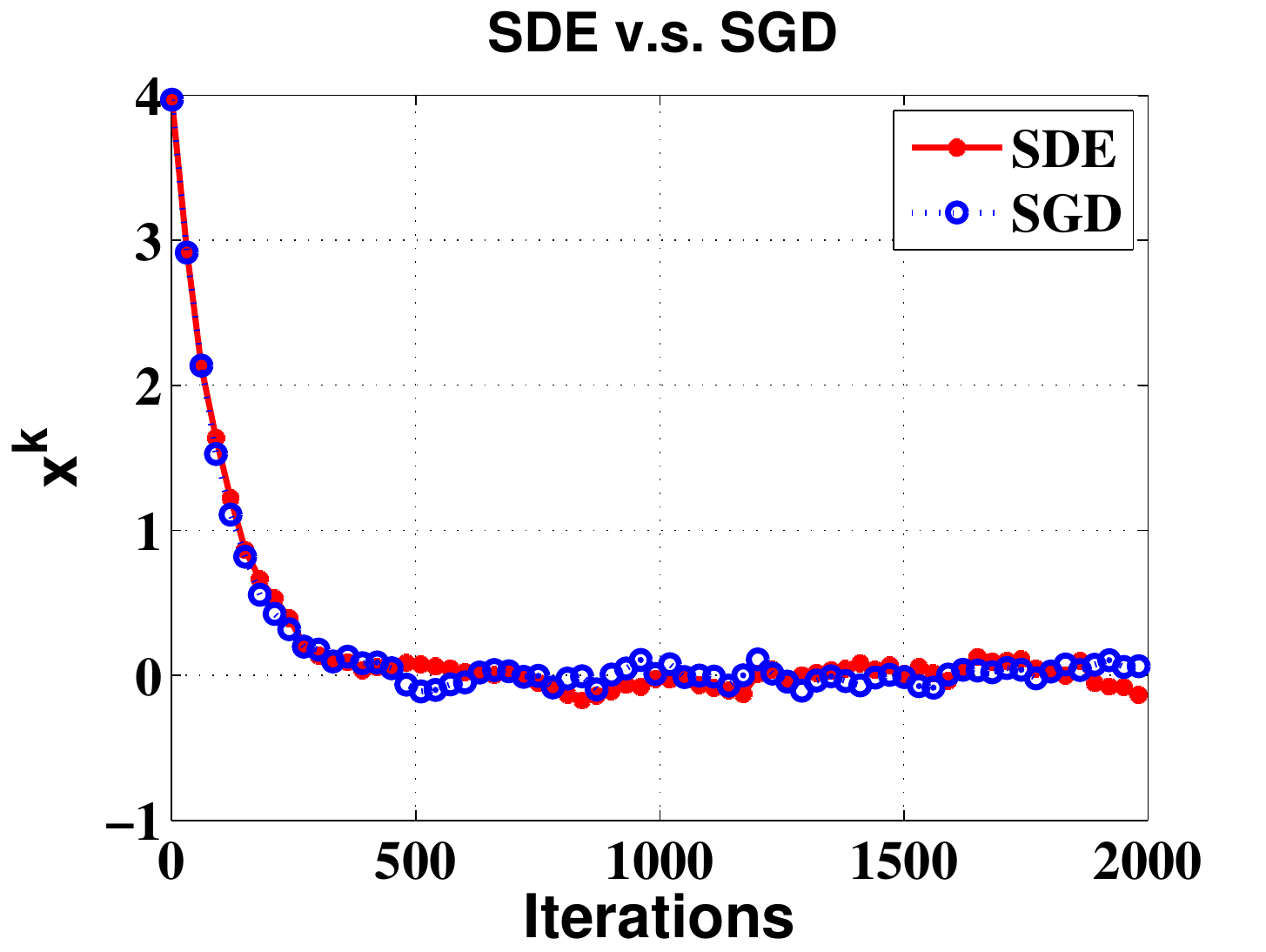}}
	\caption{Path comparison}
	\label{Fig1}
\end{figure}

For optimization iteration and Euler-Maruyama scheme, the learning rate is set to be $\eta=0.005$. We run the experiments within 2000 iterations. The figures include ASGD (resp. SGD) path and a sample path for SDDE (resp. SDE) approximation. First, we analyzed SDDE for ASGD algorithm. The time interval of the SDDE is $[0,T]$ where $T=\eta K$ and $K$ is the total number of ASGD iterations. For SDDE, we set the initial function as $\xi(\theta)=4$ for any $\theta\in [-\tau,0]$. For ASGD, we assume a constant delay as $l=10$ and let $X(0)=4$. Since the delay $l=10$, we run the SGD during the first ten steps and ASGD iterations are used from ten steps on. It is observed that the two paths are close. Second, from Fig.\ref{fig1a} we can see that the SDE approximation and SGD iteration are well matched.

\section{Conclusions}\label{con}
In this paper, we studied the continuous approximation of asynchronous stochastic gradient descent algorithm. We analyzed the approximation error and conducted theoretical analyses on the convergence rates of asynchronous stochastic gradient decent algorithms by continuous methods: moment estimation and energy function minimization. From the aspect of continuous view, we cannot only obtain existing results by the discrete view, but also get new results. For applications of this view, it can also be referred at \cite{mandt2016variational,li2016online}. To the best of our knowledge, we firstly give a continuous description to asynchronous stochastic gradient descent, and proved a tighter bound. In the future, we will apply this unified continuous view to analyze more optimization algorithms in more tasks.

\section*{Acknowledgements}
We would like to thank Liping Li and Jianhai Bao for their helpful discussions and suggestions. Zhi-Ming Ma was partially supported by National Center for Mathematics and Interdisciplinary Sciences of Chinese Academy of Sciences.


\bibliographystyle{named}
\bibliography{continuousreference}

\begin{thebibliography}{}

\bibitem[\protect\citeauthoryear{Agarwal and
  Duchi}{2011}]{agarwal2011distributed}
Alekh Agarwal and John~C Duchi.
\newblock Distributed delayed stochastic optimization.
\newblock In {\em Advances in Neural Information Processing Systems}, pages
  873--881, 2011.

\bibitem[\protect\citeauthoryear{Balles \bgroup \em et al.\egroup
  }{2017}]{Balles2016Coupling}
Lukas Balles, Javier Romero, and Philipp Hennig.
\newblock Coupling adaptive batch sizes with learning rates.
\newblock {\em Conference on Uncertainty in Artificial Intelligence}, 2017.

\bibitem[\protect\citeauthoryear{Bao \bgroup \em et al.\egroup
  }{2014}]{bao2014ergodicity}
Jianhai Bao, George Yin, and Chenggui Yuan.
\newblock Ergodicity for functional stochastic differential equations and
  applications.
\newblock {\em Nonlinear Analysis: Theory, Methods \& Applications}, 98:66--82,
  2014.

\bibitem[\protect\citeauthoryear{Bao \bgroup \em et al.\egroup
  }{2016}]{bao2016asymptotic}
Jianhai Bao, George Yin, and Chenggui Yuan.
\newblock Asymptotic analysis for functional stochastic differential equations,
  2016.

\bibitem[\protect\citeauthoryear{Dean \bgroup \em et al.\egroup
  }{2012}]{dean2012large}
Jeffrey Dean, Greg Corrado, Rajat Monga, Kai Chen, Matthieu Devin, Mark Mao,
  Andrew Senior, Paul Tucker, Ke~Yang, Quoc~V Le, et~al.
\newblock Large scale distributed deep networks.
\newblock In {\em Advances in neural information processing systems}, pages
  1223--1231, 2012.

\bibitem[\protect\citeauthoryear{Durrett}{2010}]{durrett2010probability}
Rick Durrett.
\newblock {\em Probability: theory and examples}.
\newblock Cambridge university press, 2010.

\bibitem[\protect\citeauthoryear{Johnson and
  Zhang}{2013}]{johnson2013accelerating}
Rie Johnson and Tong Zhang.
\newblock Accelerating stochastic gradient descent using predictive variance
  reduction.
\newblock In {\em Advances in Neural Information Processing Systems}, pages
  315--323, 2013.

\bibitem[\protect\citeauthoryear{Krichene \bgroup \em et al.\egroup
  }{2015}]{Krichene2015Accelerated}
W~Krichene, Am~Bayen, and Pl~Bartlett.
\newblock Accelerated mirror descent in continuous and discrete time.
\newblock pages 2845--2853, 2015.

\bibitem[\protect\citeauthoryear{Langford \bgroup \em et al.\egroup
  }{2009}]{langford2009slow}
John Langford, Alex~J Smola, and Martin Zinkevich.
\newblock Slow learners are fast.
\newblock {\em Advances in Neural Information Processing Systems},
  22:2331--2339, 2009.

\bibitem[\protect\citeauthoryear{Li \bgroup \em et al.\egroup
  }{2016}]{li2016online}
Chris~Junchi Li, Zhaoran Wang, and Han Liu.
\newblock Online ica: Understanding global dynamics of nonconvex optimization
  via diffusion processes.
\newblock In {\em Advances in Neural Information Processing Systems}, pages
  4961--4969, 2016.

\bibitem[\protect\citeauthoryear{Li \bgroup \em et al.\egroup
  }{2017}]{li2017dynamics}
Qianxiao Li, Cheng Tai, and Weinan E.
\newblock Stochastic modified equations and adaptive stochastic gradient
  algorithms.
\newblock In {\em Proceedings of the 34th International Conference on Machine
  Learning}, pages 2101--2110, 2017.

\bibitem[\protect\citeauthoryear{Lian \bgroup \em et al.\egroup
  }{2015}]{Lian2015Asynchronous}
Xiangru Lian, Yijun Huang, Yuncheng Li, and Ji~Liu.
\newblock Asynchronous parallel stochastic gradient for nonconvex optimization.
\newblock In {\em International Conference on Neural Information Processing
  Systems}, pages 2737--2745, 2015.

\bibitem[\protect\citeauthoryear{Mandt \bgroup \em et al.\egroup
  }{2016}]{mandt2016variational}
Stephan Mandt, Matthew~D Hoffman, and David~M Blei.
\newblock A variational analysis of stochastic gradient algorithms.
\newblock In {\em Proceedings of The 33rd International Conference on Machine
  Learning}, pages 354--363, 2016.

\bibitem[\protect\citeauthoryear{Mao}{2007}]{mao2007stochastic}
Xuerong Mao.
\newblock {\em Stochastic differential equations and applications}.
\newblock Elsevier, 2007.

\bibitem[\protect\citeauthoryear{Raginsky and
  Bouvrie}{2012}]{Raginsky2012Continuous}
Maxim Raginsky and J~Bouvrie.
\newblock Continuous-time stochastic mirror descent on a network: Variance
  reduction, consensus, convergence.
\newblock 23(1):6793--6800, 2012.

\bibitem[\protect\citeauthoryear{Rakhlin \bgroup \em et al.\egroup
  }{2011}]{rakhlin2011making}
Alexander Rakhlin, Ohad Shamir, and Karthik Sridharan.
\newblock Making gradient descent optimal for strongly convex stochastic
  optimization.
\newblock {\em arXiv preprint arXiv:1109.5647}, 2011.

\bibitem[\protect\citeauthoryear{Recht \bgroup \em et al.\egroup
  }{2011}]{recht2011hogwild}
Benjamin Recht, Christopher Re, Stephen Wright, and Feng Niu.
\newblock Hogwild: A lock-free approach to parallelizing stochastic gradient
  descent.
\newblock In {\em Advances in Neural Information Processing Systems}, pages
  693--701, 2011.

\bibitem[\protect\citeauthoryear{Su \bgroup \em et al.\egroup
  }{2014}]{su2014differential}
Weijie Su, Stephen Boyd, and Emmanuel Candes.
\newblock A differential equation for modeling nesterov's accelerated gradient
  method: theory and insights.
\newblock In {\em Advances in Neural Information Processing Systems}, pages
  2510--2518, 2014.

\bibitem[\protect\citeauthoryear{Uhlenbeck and
  Ornstein}{1930}]{uhlenbeck1930theory}
George~E Uhlenbeck and Leonard~S Ornstein.
\newblock On the theory of the brownian motion.
\newblock {\em Physical review}, 36(5):823, 1930.

\bibitem[\protect\citeauthoryear{Wibisono \bgroup \em et al.\egroup
  }{2016}]{wibisono2016variational}
Andre Wibisono, Ashia~C Wilson, and Michael~I Jordan.
\newblock A variational perspective on accelerated methods in optimization.
\newblock {\em Proceedings of the National Academy of Sciences}, page
  201614734, 2016.

\bibitem[\protect\citeauthoryear{Yang \bgroup \em et al.\egroup
  }{2017}]{Yang2017The}
Lin~F. Yang, R.~Arora, V.~Braverman, and Tuo Zhao.
\newblock The physical systems behind optimization algorithms.
\newblock 2017.

\bibitem[\protect\citeauthoryear{Zhang \bgroup \em et al.\egroup
  }{2015}]{zhang2015deep}
Sixin Zhang, Anna~E Choromanska, and Yann LeCun.
\newblock Deep learning with elastic averaging sgd.
\newblock In {\em Advances in Neural Information Processing Systems}, pages
  685--693, 2015.

\bibitem[\protect\citeauthoryear{Zheng \bgroup \em et al.\egroup
  }{2017}]{zheng2016asynchronous}
Shuxin Zheng, Qi~Meng, Taifeng Wang, Wei Chen, Nenghai Yu, Zhi-Ming Ma, and
  Tie-Yan Liu.
\newblock Asynchronous stochastic gradient descent with delay compensation.
\newblock In {\em International Conference on Machine Learning}, pages
  4120--4129, 2017.

\end{thebibliography}

\end{document}